\documentclass[conference]{IEEEtran}

\usepackage{hyperref}
\usepackage{graphicx}
\usepackage{listings}
\usepackage[normalem]{ulem}

\ifCLASSINFOpdf
\else
\fi
\begin{document}
\title{Cybonto: Towards Human Cognitive Digital Twins for Cybersecurity}

\author{\IEEEauthorblockN{Tam n. Nguyen}
\IEEEauthorblockA{\textit{Department of Management Information Systems}\\University of Arizona\\tamn@email.arizona.edu\\linkedin.com/in/tamcs/}
}

\maketitle

\begin{abstract}
Cyber defense is reactive and slow. On average, the time-to-remedy is hundreds of times larger than the time-to-compromise. In response to the expanding ever-more-complex threat landscape, Digital Twins (DTs) and particularly Human Digital Twins (HDTs) offer the capability of running massive simulations across multiple knowledge domains. Simulated results may offer insights into adversaries' behaviors and tactics, resulting in better proactive cyber-defense strategies. For the first time, this paper solidifies the vision of DTs and HDTs for cybersecurity via the Cybonto conceptual framework proposal. The paper also contributes the Cybonto ontology, formally documenting 108 constructs and thousands of cognitive-related paths based on 20 time-tested psychology theories. Finally, the paper applied 20 network centrality algorithms in analyzing the 108 constructs. The identified top 10 constructs call for extensions of current digital cognitive architectures in preparation for the DT future.
\end{abstract}


\IEEEpeerreviewmaketitle

\section{Introduction}
Humans are recognized to be the weakest link in the cybersecurity defense chain \cite{Bulgurcu2010InformationAwareness, MaalemLahcen2020ReviewCybersecurity}. Insider threat incidents cost both small and large companies billions of dollars annually \cite{PonemonInstitute20202020Report}. Cyber defenders are reactive and slow. On average, hackers need 15 hours to compromise a system while defenders need 200 to 300 days to discover a breach \cite{MaalemLahcen2020ReviewCybersecurity}. Meanwhile, the cybersecurity threat landscape keeps expanding. Cyber defenders respond by enlisting inter-discipline knowledge from numerous fields such as math, psychology, and criminology \cite{Carley2020SocialScience, Li2020Theoryon:Learning, Valja2020AutomatingInfrastructures, MaalemLahcen2020ReviewCybersecurity}. In such a climate, Digital Twins (DTs) and incredibly Human Digital Twins (HDTs) offer the capability of running large-scale simulations across multiple knowledge domains to improve proactive cyber-defense strategies. 

Digital Twins are computational models of physical systems, including humans. The DT market is rapidly growing at a compound annual rate of 45.4\% \cite{Eirinakis2020EnhancingTwins}. Notably, massive DT projects such as the British National Digital Twin \cite{centre-for-digital-built-britain-2019} are being built. Within the intertwined DT networks, individual smart DTs such as HDTs should be capable of not only executing mimetic behaviors but also having local and global awareness, self-learning, and self-optimizing \cite{Eirinakis2020EnhancingTwins}.

For the first time, this paper proposes a grounded vision on how DTs and HDTs can be applied towards cybersecurity. The main goal is making a case for expanding current digital cognitive architectures that will be at the hearts of future HDTs. The paper unified twenty most cybersecurity-relevant finalists from over seventy behavioral psychology theories. The theory-informed knowledge and other cybersecurity constructs were then encoded as the novel Cybonto ontology. Analyzing the Cybonto ontology informed the Cybonto conceptual framework.

The key contributions are as followed. The Cybonto conceptual framework solidifies the vision of how human cognitive digital twins and digital twin systems can be leveraged to design proactive cybersecurity strategies. The Cybonto ontology provides research-based guidance on 108 constructs and thousands of possible paths among them. Analyzing the ontology's cognitive core using more than 20 network centrality algorithms yields Behavior, Arousal, Goals, Perception, Self-efficacy, Circumstances, Evaluating, Behavior-Controlabiity, Knowledge, and Intentional Modality as the top 10 most influential constructs. These results call for the expansion of current cognitive architectures to better fit their future employments in DT systems.
%\hfill tnn
 %\hfill August 04, 2021

\section{Literature Review}
The concept of HDTs previously appeared in human-computer interaction studies. In comparison with traditional models, HDTs for digital twin systems have broader scopes with emphasis on both behavioral and cognitive activities. The work of Somers et al. is an excellent example in which HDT acts as a sensible personal assistant in organizing social events \cite{Somers2020CognitiveAssistants}. Notably, the HDT did not explicitly ask potential event participants for their preferences. Instead, it observed the people's social dimensions and then modeled the cognitive processes underlying an expert event planner's decisions.

Zhang et al. \cite{Zhang2020TowardsSelf-Awareness} describes HDTs' self-awareness as a continuous process that involves dynamic knowledge acquisition and utilization. Numerous feedback loops will be needed. Well-designed ontologies are essential for information exchanges among different models \cite{Ma2019TheReview, Bienvenu2014Ontology-basedMMSNP}. Compared with an application ontology, a reference ontology is supposed to be much more canonical and reusable \cite{Arp2015BuildingOntology}. Ontological reusability begins with the adoption of a top-level ontology. Key papers in cognitive frameworks and cybersecurity ontologies are as followed.

\subsection{Cognitive Frameworks}

ACT-R \cite{Anderson1997ACT-R:Attention} is representative of the psychological modeling group with Clarion and Epic as other members. SOAR \cite{Laird2019TheArchitecture} is representative of the agent functionality-focused group that also includes Sigma, Lida, Icarus, and Companions. ACT-R and SOAR differ on architectural constraints, memory retrieval, conflict resolution strategies, and exhaustive processing \cite{Jones2007ComparingSOAR}. ACT-R sequential architecture forces developers to watch out for bottlenecks while SOAR's parallel architecture is more relaxed \cite{Jones2007ComparingSOAR}. ACT-R provides two options for resolving conflicts, while SOAR offers none.

Both SOAR and ACT-R share the same general cognitive cycle and common architectural modules such as perception, short-term memory, declarative learning, declarative long-term memory, procedural long-term memory, procedural learning, action selection, and action. While ACT-R, SOAR, and other cognitive systems rely on symbolic input/output and rule database, their symbols may contain statistical metadata, and their architectures do allow the integration of deep learning systems.

\subsection{Cybersecurity Ontologies}
Ontologies are essential for symbolic operations, the building of a knowledge base, and explainability. Ontologies can be manually build from scratch \cite{Uschold1995TowardsOntologies, Uschold1995TheOntology} or be automatically extracted \cite{Maedche2003ManagingWeb, Haase2005ConsistentOntologies}. DOLCE\footnote{\url{https://lnkd.in/gTFR8Wt}} vs. BFO\footnote{\url{https://basic-formal-ontology.org/}} highlights the importance of ontological commitments. DOLCE is grounded in natural language while BFO is grounded in the real world \cite{Partridge2020AModel}. Because objects can be conceptual or actual in a language-based ontology, there is always a risk of one actual object being recognized as two or more different conceptual objects.

Oltramari et al. \cite{Oltramari2014BuildingSecurity} introduced Cratelo, which has DOLCE as the top-level ontology. The ontology's human behavioral structures are confined within the cyber operation scope. Costa et al. \cite{Costa2016AnOntology} used the natural language processing approach in building their Insider Threat Indicator Ontology (ITIO). The ontology inherited considerable amounts of language ambiguity and did not support the identification of deeper behavioral structures. In 2019, Greitzer et al. \cite{Greitzer2019DesignOntology} built upon their 2016's work and introduced the Sociotechnical and Organizational Factors for Insider Threat (SOFIT). Due to the absence of a top-level ontology and the behavioral language that leans heavily towards organizational insider threat activities, SOFIT is an application ontology rather than a reference ontology. Greitzer et al. \cite{Greitzer2019DesignOntology} also admitted that ontology validation exercises only covered 10\% of the ontology.

Meanwhile, Donalds and Osei-Bryson \cite{Donalds2019TowardApproachb} reported that cybersecurity ontologies have been insufficient due to fragmentation, incompatibility, and inconsistent use of terminologies. The team proposed a cybercrime classification ontology structured around attack events \cite{Donalds2019TowardApproachb}. While the ontology provides a holistic, multi-perspective view regarding cybercrime attacks, its behavioral components are limited and lack theoretical grounding.

\begin{table*}
\renewcommand{\arraystretch}{1.2}
\caption{Ranking of Top 25 Cybersecurity Related Behavioral Theories}
\label{tab:theoryRanking}
\centering
\begin{tabular}{|l|l|l|l|l|l|l|l|l|l|l|}
\hline
Code & \textbf{Theory Name} & Year & Citations & \begin{tabular}[c]{@{}l@{}}Author\\ i-10\end{tabular} & { \begin{tabular}[c]{@{}l@{}}G-scholar\\ Results\end{tabular}} & \begin{tabular}[c]{@{}l@{}}Cyber-\\ security\\ Impres-\\ sions\end{tabular} & \begin{tabular}[c]{@{}l@{}}Cyber-\\ security\\ Density\end{tabular} & \begin{tabular}[c]{@{}l@{}}Crimi-\\ nology\\ Impres-\\ sions\end{tabular} & \begin{tabular}[c]{@{}l@{}}Fitted\\ Cita-\\ tions\\ /Year\end{tabular} & \textbf{\begin{tabular}[c]{@{}l@{}}Final\\ Score\end{tabular}} \\ \hline
28 & { \textbf{Protection Motivation Theory}} & 1997 & 1211 & n/a & { 10500} & 10 & 9 & 7 & 0 & \textbf{6.5} \\ \hline
27 & { \textbf{Prospect Theory}} & 2013 & 68709 & 354 & { 66200} & 8 & 1 & 6 & 10 & \textbf{6.3} \\ \hline
15 & { \textbf{General Theory of Crime}} & 1990 & 14412 & 91 & { 13500} & 9 & 1 & 10 & 1 & \textbf{5.3} \\ \hline
35 & { \textbf{Self-Efficacy Theory}} & 1999 & 91578 & n/a & { 212000} & 9 & 0 & 6 & 5 & \textbf{5} \\ \hline
44 & { \textbf{Social Norms Theory}} & 2005 & 534 & n/a & { 47400} & 7 & 9 & 2 & 0 & \textbf{4.5} \\ \hline
2 & { \textbf{Affective Events Theory}} & 1996 & 5587 & 66 & { 6880} & 10 & 1 & 6 & 0 & \textbf{4.3} \\ \hline
10 & { \textbf{Differential Association   Theory}} & 1995 & 148 & 31 & { 10700} & 9 & 1 & 7 & 0 & \textbf{4.3} \\ \hline
12 & { \textbf{Extended Parallel Processing Model}} & 1992 & 3453 & n/a & { 412} & 7 & 4 & 6 & 0 & \textbf{4.3} \\ \hline
14 & { \textbf{Focus Theory of Normative Conduct}} & 1991 & 3116 & n/a & { 6220} & 6 & 10 & 1 & 0 & \textbf{4.3} \\ \hline
8 & { \textbf{Containment Theory}} & 1961 & 638 & 43 & { 2240} & 9 & 1 & 6 & 0 & \textbf{4} \\ \hline
49 & { \textbf{Theory of Planned Behaviour}} & 1991 & 88719 & 217 & { 85800} & 9 & 1 & 3 & 3 & \textbf{4} \\ \hline
41 & { \textbf{Social Identity Theory}} & 1979 & 25410 & n/a & { 66200} & 7 & 0 & 7 & 1 & \textbf{3.8} \\ \hline
18 & { \textbf{Goal Setting Theory}} & 1990 & 14489 & 294 & { 51700} & 6 & 1 & 7 & 1 & \textbf{3.8} \\ \hline
51 & { \textbf{\begin{tabular}[c]{@{}l@{}}Transtheoretical Model of  \\ Behaviour Change\end{tabular}}} & 1997 & 7842 & 307 & { 35900} & 6 & 0 & 7 & 0 & \textbf{3.3} \\ \hline
34 & { \textbf{Self-Determination Theory}} & 2012 & 1879 & 319 & { 165000} & 8 & 0 & 4 & 0 & \textbf{3} \\ \hline
22 & { \textbf{Operant Learning Theory}} & 1965 & 22236 & 461 & { 40500} & 7 & 1 & 4 & 0 & \textbf{3} \\ \hline
39 & { \textbf{Social Cognitive Theory}} & 2001 & 21385 & n/a & { 162000} & 8 & 0 & 3 & 1 & \textbf{3} \\ \hline
5 & { \textbf{Change Theory}} & 1958 & 74 & n/a & { 54700} & 8 & 0 & 2 & 0 & \textbf{2.5} \\ \hline
23 & { \textbf{\begin{tabular}[c]{@{}l@{}}Precaution Adoption Process  \\ Approach\end{tabular}}} & 1988 & 1804 & 102 & { 2590} & 6 & 1 & 3 & 0 & \textbf{2.5} \\ \hline
11 & { \textbf{Diffusion of Innovations}} & 1962 & 125502 & n/a & { 96700} & 4 & 1 & 3 & 2 & \textbf{2.5} \\ \hline
9 & { \textbf{Control Theory}} & 1982 & 3113 & 381 & { 11500} & 6 & 1 & 1 & 0 & \textbf{2} \\ \hline
33 & { \textbf{Risk as Feelings Theory}} & 2001 & 6796 & 313 & { 550} & 5 & 2 & 1 & 0 & \textbf{2} \\ \hline
43 & { \textbf{Social Learning Theory}} & 1941 & 3761 & n/a & { 145000} & 2 & 0 & 6 & 0 & \textbf{2} \\ \hline
21 & { \textbf{Norm Activation Theory}} & 1992 & 18807 & 234 & { 4610} & 5 & 1 & 1 & 1 & \textbf{2} \\ \hline
45 & { \textbf{Technology Acceptance Model}} & 2003 & 33231 & 123 & { 48100} & 2 & 3 & 1 & 2 & \textbf{2} \\ \hline
\end{tabular}
\end{table*}

\begin{figure*}[!t]
\centering
\includegraphics[width=6.5in]{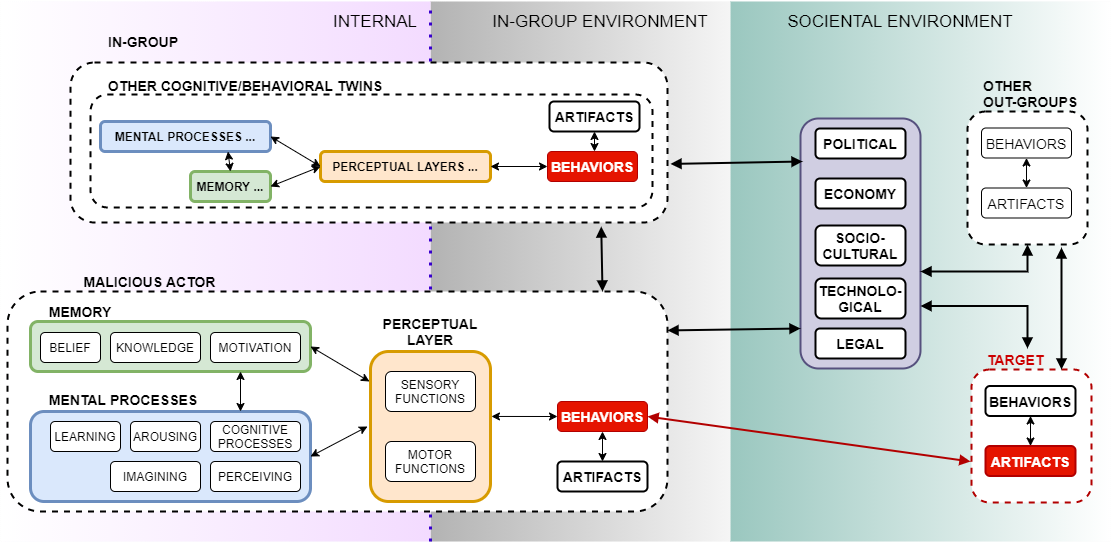}
\caption{Conbonto Conceptual Framework}
\label{fig_cybonto_concept}
\end{figure*}

\subsection{Open Problems}
While massive DT projects are underway, digital cognitive twin development is pale in comparison, and HDT for cybersecurity is non-existent. This paper examined both ACT-R and SOAR published research repositories and found no cybersecurity dedicated track with topics such as cybersecurity, online ethical decisions, cyber criminology, or cyber attack/defense simulations. There is no grounded vision on how powerful DT systems with HDTs may improve proactive cybersecurity defenses. Recommended exploring questions are (i) What are the types of HDT (malicious hackers, groups as single HDT, defenders,etc.) to be built? (ii) What will HDT for cybersecurity feedback loops look like? (iii) How will existing cognitive architectures be extended to best facilitate those feedback loops? (iv) What shall we learn from our continuous observation of those HDTs? 

\begin{figure*}[!t]
\centering
\includegraphics[width=7in]{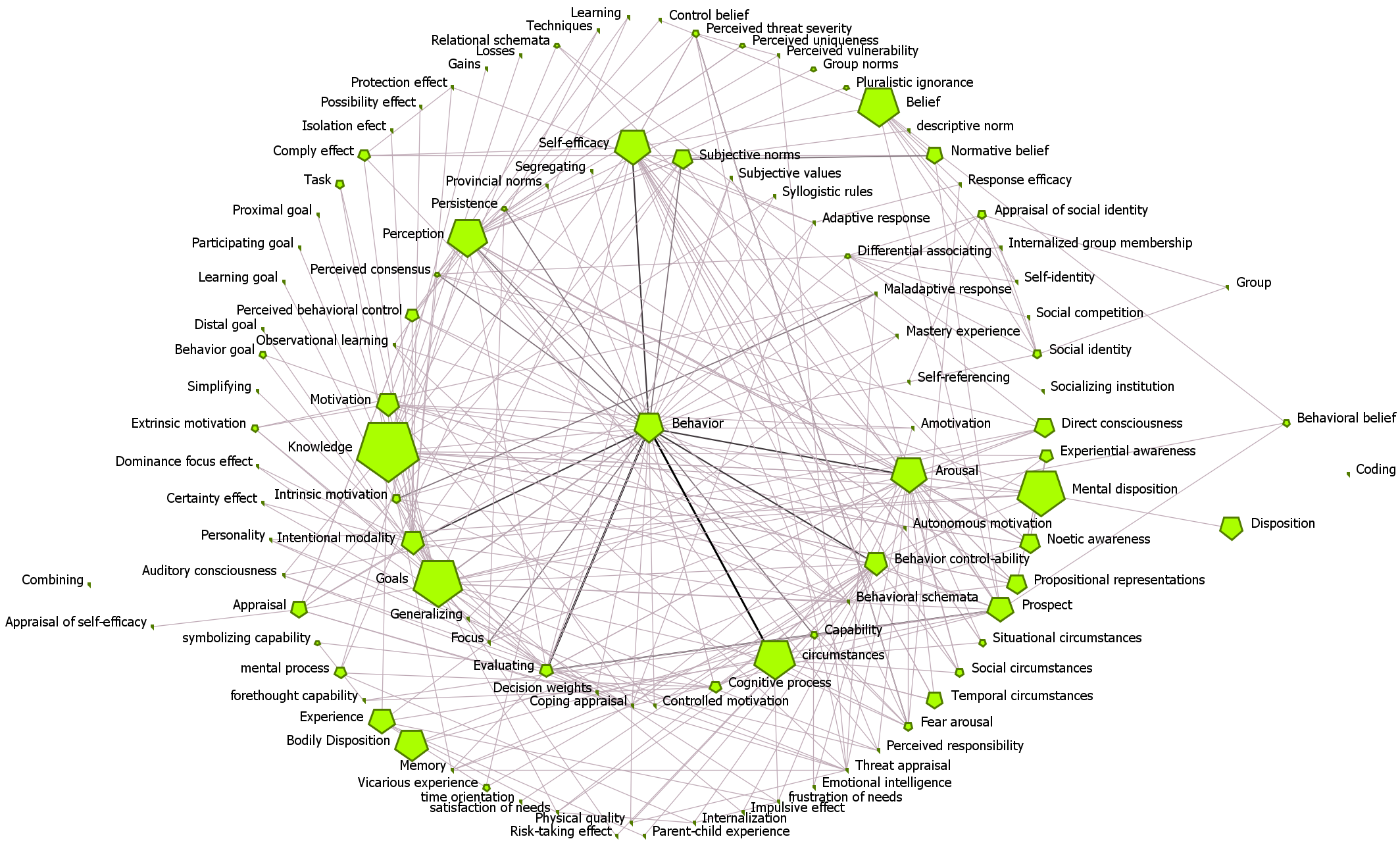}
\caption{Conbonto Ontology - Behavioral/Cognitive's Horizontal Relationships Visualized}
\label{fig_HoriRelationships}
\end{figure*}

Current cybersecurity-related cognitive models focus on narrow use cases and are far from the HDTs that can automatically interact with other DTs while building up their own awareness. For the main reason, existing cognitive architectures do not provide enough granularity. This leads to further problems with multi-modal understanding and meta-cognition. For example, current long-term memory architecture can be further divided into experiences and beliefs. It is possible for two persons sharing a strong belief to have different interpretations of the same data. One may be significantly influenced by a past experience. Additionally, having access to too much data due to lack of granularity will lead to cognitive bottlenecks at system levels. Deciding which chunks of data to be loaded, excluded or be permanently erased from memory remains a challenge.

Finally, we do not have a reference ontology for documenting and sharing behavioral-cybersecurity knowledge. Existing cybersecurity ontologies that have behavioral components are mostly application ontologies with none or weak ontological commitments. Such ontologies will not fit for use in massive and complex DT systems. 

\section{The Cybonto Conceptual Framework}
The novel Cybonto conceptual framework aims to provide general directions on answering the previously-mentioned questions regarding the vision of DTs and HDTs for cybersecurity. The framework targets the cognitive process of a malicious actor as an HDT within a DT system. Cognitive space is defined by the behavioral/cognitive component of the Cybonto ontology. The action space is limited by the HDT's set of encoded actions, its ability to improvise new moves, and the other DTs' interaction interfaces. In the beginning, fifty theories were picked from the behavioral/cognitive psychology body of knowledge. Each theory was ranked based on its ability to generate research, relevancy to cybersecurity and criminology, and consistency. Table \ref{tab:theoryRanking} presents the top 25 theories. 

Then, each theory was codified into tuples of (entity, "influence" relationship, entity). The combination of 20 codified top theories formed the Cybonto cognitive core ontology with over 100 constructs. Full description of Cybonto in RDF store, Neo4J relational database, and other documentation are available at Cybonto-1.0 Github repository \footnote{https://github.com/Cybonto/CYBONTO-1.0}. The Cybonto conceptual framework was formed upon analysis of the Cybonto ontology. Figure \ref{fig_cybonto_concept} presents the Cybonto conceptual framework with three environment types and four groups of digital twins (DTs). 

The internal environment (INE) is private to each DT. It contains both cognitive components and non-cognitive components. Opposite to the internal environment is the societal environment (SOE) where everything is public. In between, the in-group environment (IGE) connects INE with SOE. All environments follow Bronfenbrenner's Ecological System Theory \cite{Bronfenbrenner1992EcologicalTheory.} which describes influences as progressive, varying, and reciprocal forces among individuals and environments. For example, a seemingly distant public event may still be able to affect certain private mental processes. 

The IEG and the SOE are relative to the targeted HDT. The IEG is equivalent to Bronfenbrenner's micro- and meso-systems. The microsystem is the most inﬂuential external environment with members such as family, close friends, school, lovers, and mentors. SOE is equivalent to Bronfenbrenner's Exo-, Macro-, and Chrono-systems. The Cybonto conceptual framework requires four representatives from four DT groups. We need one attacker HDT and one defender HDT. Unlike traditional models to which data and feature specifications were explicitly fed, an attacker HDT must collect the data by itself. Group-related data cannot be inferred if the fundamental group structure is not met. Hence, we then need at least two more DTs to present IEG and SOE identities.

An HDT can perform two main types of behaviors: the artifact creating/altering behavior and the non-artifact behavior. An artifact can range from a piece of code to a complex non-cognitive digital twin. Viewing a malware's codes is a non-artifact behavior, while running the codes can be an artifact-altering behavior if the codes make changes to other artifacts. The perceptual layer sits on the border between the internal and external environments (IEG and SOE). Different perceptual layers in combination with different cognitive systems will have different perceptions of the same data streams. Refined perceptions constitute only a small part of a digital cognitive system. The Cybonto ontology details thousands of cognitive paths for processing initial perceptions. The result of a cognitive processing chain will be either a non-artifact behavior or an artifact creating/altering behavior. The behaviors (data streams) will be observed by other HDTs, and a new round of feedback loops begins. It is essential to note that a behavior can be kept secret within the in-group environment.

In this framework: (a) HDTs have the complete freedom to interact with other DTs per published protocols, and automatically seek whatever data is made available to them. (b) By releasing their behaviors, HDTs generate new data, which may then be consumed by other HDTs. (c) The cognitive architecture within each HDT determines its cognitive capabilities which should include awareness and adaptation. (d) Cybonto DT simulation's objectives should be more about discovering new knowledge (the Why and How) rather than mining specific data (the What).

\section{Analysis of The Cybonto Ontology}
% PENDING TASKS: add more object properties (hasInitTime, hasExpirationTime, hasBehaviorChainID, hasSessionID, hasPeriodID)

Cybonto elected the Basic Formal Ontology (BFO) as its top-level ontology from more than thirty candidates. BFO is the only top-level ontology that adopts materialism, commits to actual-world possibilia, and has an intensional criterion of identity. The Cybonto Core (the behavioral/cognitive component) is grounded further by employing the Mental Functioning (MF) as its mid-level ontology. MF follows best practices outlined by the OBO Foundry and aligns with other projects in the Cognitive Atlas - a state-of-the-art collaborative knowledge-base in Cognitive Science \cite{Hastings2012RepresentingDisease}.

Materialism views the world as a collection of materialized objects existing in space and time \cite{Partridge2020AModel} - a core principle in DT strategies. Committing to materialism through BFO offers a fundamental distinction in the way Cybonto represents mental constructs. For centuries, cognitive activities were considered abstract particulars that could only be described through languages. This tradition is the reason why most behavioral components in cybersecurity ontologies are language-based. Recent breakthroughs in the brain-machine interface such as those of Neuralink \cite{Musk2019AnChannels} enables measurements of brain activities that correspond to certain cognitive constructs. Therefore, it is now possible to ground behavioral/cognitive ontologies in materialism. Cybonto rejects conceptual objects, different linguistic descriptions of the same actual objects, process-based objects, and qualitative object labels that cannot be measured in real life.

\begin{figure}[!t]
\centering
\includegraphics[width=3.45in]{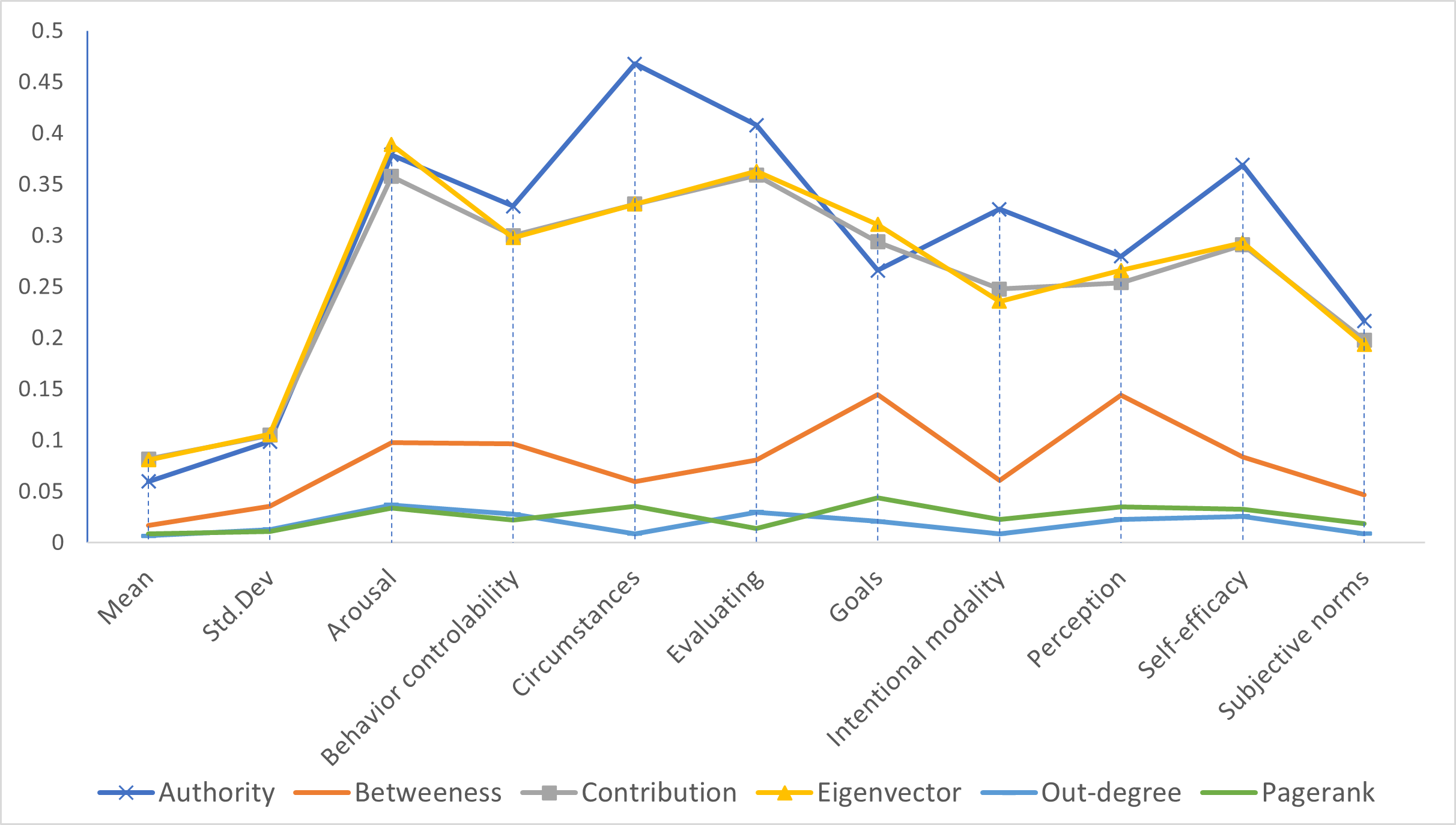}
\caption{Most Influential Constructs Based on Six Types of Centralities}
\label{fig_TopEntities}
\end{figure}

Figure \ref{fig_HoriRelationships} shows the network of Cybonto's horizontal relationships. Each node size equals the log scale of the node's page rank. A darker link color indicates a higher link value. Nodes were automatically arranged in a multi-circle layout with higher betweenness centrality nodes closer to the center. Figure \ref{fig_TopEntities} shows the most popular entities based on different network centrality scores.

Top Authority Central (AC) constructs receive influence from constructs that have the most influence on others. Top Betweenness Central (BC) constructs are the ones that sit in the shortest paths among other constructs. BC constructs can serve either as bridges or gatekeepers of other constructs and processes. Top Eigenvector Central (EC) constructs are the leaders of their own cliques. A clique is a group of constructs in which each member has relationships with the others. In the context of the cognitive digital twin, a clique may represent a strong cognitive/behavioral pattern. Not only the top EC constructs are well-connected with their own clique members, they also have relationships with other cliques.

Contribution Centrality is Eigenvector Centrality on inverse-Jaccard weighted values of the input networks. A link between two constructs has the most contribution weight when the neighbors of one end are most different from the neighbors at the other end. Top Out-degree Central (OC) constructs have the most out-links (influencing) to others while top In-coming Central (IC) constructs are influenced by the most important in-coming neighbors. The top PageRank constructs have relationships (whether in-coming or out-going) with the most influential neighbors. 
% summary of centralities - https://arxiv.org/pdf/2011.14575.pdf

The top 10 constructs across 20 network centrality measures are Behavior, Arousal, Goals, Perception, Self-efficacy, Circumstances, Evaluating, Behavior-Controlabiity, Knowledge, and Intentional Modality. In this list, only Behaviors, Goals, Perception, Evaluating and Knowledge are parts of existing digital cognitive architectures, although some are not explicitly implemented. It is possible that before this study, influential cognitive structures have been studied per independent use-cases and thus could not collectively attract attention from conservative cognitive system designers. Now with a bird-eye view across 20 behavioral theories, these top 10 constructs deserve better attention.

Within cognitive architectures, we may consider implementing Goals, Knowledge, Perception, and Evaluating explicitly and with finer granularity. For example, Perception is more than short-lived sensory perception. For example, Alice perceived Bob as a nice guy, and such perception persists whether Bob is with Alice or not. Additionally, we should consider adding Arousal and Intentional Modality. Although Arousal is a non-cognitive construct, it is ranked in the second place and influences several cognitive constructs within the top 10, such as Evaluating and Intentional Modality. Unfortunately, the current state of research regarding Arousal as a part of a digital cognitive process is almost non-existent. SOAR-related research results show a few papers studying the effects of general emotions. ACT-R research repository shows just four papers studying the effects of Arousal on memory management.

The Circumstance is another non-cognitive construct with significant influence on behavioral outcomes. The paper recommends expanding the existing Mental Image module in existing cognitive architectures to include non-physical environment variables such as urgency, group dynamics, and social sentiments. Finally, the paper recommends a new component - Imagining - to enable the HDT to run its own situational simulations and reason about possible circumstances.

\section{Conclusion}
Once massive non-cognitive digital twin systems are brought online, adding human cognitive digital twins will be the only logical next step. The vision of letting human digital twins run free in a digital twin world (and observing them) is realistic and offers a new paradigm in knowledge mining. The Cybonto conceptual framework demonstrates how such an ecosystem can be leveraged for shaping proactive cybersecurity defense strategies. Notably, HDTs are fundamentally different from deep learning models. Most cognitive systems can combine human cognitive reasoning (symbolic) with deep learning models (sub-symbolic). Cognitive reasoning with good enough granularity and a well-designed ontology allows us to observe and more importantly, to understand what the digital twins are doing. Hence, the paper also proposes the Cybonto ontology as specific recommendations on how existing cognitive systems can be expanded. Future work may involve further framework development, fine-tuning and expanding the ontology, and building a malicious HDT for demonstration purposes.

%\section*{Acknowledgment}

\bibliographystyle{IEEEtran}
\bibliography{IEEEabrv,references.bib}

\end{document}